\documentclass[journal]{IEEEtran}
 \usepackage{multirow}
 \usepackage{graphicx,cite}
 \usepackage{epstopdf}
 \usepackage{amssymb,amsmath,bm, footmisc}
 \usepackage{textcomp}
 \usepackage{algorithm}
 \usepackage{algorithmic}
\usepackage{caption} 
\usepackage{subcaption}
\usepackage{listings}

\usepackage{lettrine}
\usepackage[absolute]{textpos}
\usepackage{fancyhdr}
\usepackage{rotating}
\usepackage{pdflscape}
\usepackage{blindtext}
\usepackage{ctable}   
\newcommand{\norm}[1]{\left\lVert#1\right\rVert}

\ifCLASSINFOpdf
\else
\fi
\begin{document}
%
\title{Correlation and Class Based Block Formation for Improved Structured Dictionary Learning}
%
%
%


\author{Nagendra~Kumar\IEEEmembership{}
	and~Rohit~Sinha,~\IEEEmembership{Member,~IEEE}
	
	\IEEEcompsocitemizethanks{\IEEEcompsocthanksitem The authors are with the Department
		of Electronics and Electrical Engineering, Indian Institute of Technology Guwahati, Assam,
		India, 781039.\protect\\
		E-mail: \{k.nagendra, rsinha\}@iitg.ernet.in
		}
	}
\markboth{Submitted to  IEEE Transactions on Signal Processing}{}
\maketitle

\begin{abstract}
In recent years, the creation of block-structured dictionary has attracted a lot of interest. Learning such dictionaries involve two step process: block formation and dictionary update. Both these steps are important in producing an effective dictionary. The existing works mostly assume that the block structure is known \emph{a priori} while learning the dictionary. For finding the unknown block structure given a dictionary commonly sparse agglomerative clustering (SAC) is used. It groups atoms based on their consistency in sparse coding with respect to the unstructured dictionary. This paper explores two innovations towards improving the reconstruction as well as the classification ability achieved with block-structured dictionary. First, we propose a novel block structuring approach that makes use of the correlation among dictionary atoms. Unlike the SAC approach, which groups diverse atoms, in the proposed approach the  blocks are formed by grouping the top most correlated atoms in the dictionary. The proposed block clustering approach is noted to yield significant reductions in redundancy as well as provides a direct control on the block size when compared with the existing SAC-based block structuring. Later, motivated by works using supervised \emph{a priori} known block structure, we also explore the incorporation of class information in the proposed block formation approach to further enhance the classification ability of the block dictionary. For assessment of the reconstruction ability with proposed innovations is done on synthetic data while the classification ability has been evaluated in large variability speaker verification task.

\end{abstract}

\begin{IEEEkeywords}
Block-KSVD dictionary, block orthogonal matching pursuit, sparse representation classification.
\end{IEEEkeywords}

%
\IEEEpeerreviewmaketitle

\section{Introduction}
\lettrine{L}{earned} dictionary based sparse representation (SR) finds successful application in various signal processing domains, such as image denoising~\cite{elad2006image}, image recognition~\cite{gao2014learning,chi2013block}, face recognition~\cite{wright2009robust,patel2012dictionary,dong2016orthonormal}, speaker identification/verification~\cite{naseem2010sparse,kua2011speaker,haris2015robust}, and fingerprint identification\cite{liu2015latent}.
In SR domain, existing data driven dictionary learning techniques can be  broadly divided into three categories: supervised, semi-supervised and unsupervised. The dictionary learned utilizing class labels are referred to as supervised. Whereas those learned using weak supervision in form of any assumed structure/constraint are termed as semi-supervised dictionary. Both these kinds of dictionaries produce more discriminative sparse codes than the unsupervised ones, thus result in better classification performance. In SR domain, usually redundant (over-complete) dictionaries are preferred. Such dictionaries have more columns (atoms) than rows (data dimensionality). Sometimes, lesser number of examples than the data dimensionality involved are available for learning the dictionary. Thus, only under-complete dictionary could be learned unless we project the data to appropriate low-dimensional space. Nevertheless, the use of under-complete dictionaries have been reported in SR based classification tasks~\cite{haris2015robust, singh2013language}.

In recent past, the dictionary learning has received a lot of attention in SR domain. Combining the K-means clustering and the singular value decomposition (SVD), a widely used dictionary learning approach is proposed and is referred to as KSVD~\cite{ksvd_2006}. In learning of the KSVD dictionary, the reconstruction error is minimized under the constraint on sparsity for the given data. Though not being optimized for producing discriminative sparse codes, some works have explored the KSVD dictionaries in the classification task~\cite{wright2009robust,haris2015robust}. In the literature, a few classification-driven dictionaries are also proposed such as supervised KSVD (S-KSVD)~\cite{rodriguez2008sparse} dictionary and label-consistent KSVD (LC-KSVD)~\cite{lcksvd_2013} dictionary. The S-KSVD algorithm incorporates the Fisher discriminant criterion in dictionary learning. Whereas in the LC-KSVD algorithm, a linear transformation that maps the sparse code to more discriminative one is also learned along with the dictionary. Though yielding enhanced classification performance, these supervised dictionaries neither use any block structure nor explicitly minimize the within-class redundancy. As the result of that, such dictionaries are found to yield inconsistent sparse codes for the same class data.  

In addition to supervised dictionary, some block-structured dictionaries are also proposed. The introduction of block structure in a dictionary is noted to enhance not only its reconstruction ability~\cite{zelnik2012dictionary} but also its classification ability~\cite{sreeram2015improved}. Initial works simply exploit the known block structures in sparse coding with no emphasis on learning such dictionaries~\cite{stojnic2008reconstruction,eldar2009block,eldar2009robust,eldar2010average}. The block-KSVD (BKSVD)~\cite{zelnik2012dictionary} dictionary is probably the first attempt towards learning an unsupervised block-structured dictionary. It employs a sparse agglomerative clustering (SAC) algorithm for estimating the unknown block structure. Given a dictionary, the SAC algorithm estimates the block structure by iteratively grouping its atoms based on sparse coding. As the SAC employs orthogonal matching pursuit (OMP)~\cite{rubinstein2008efficient} for sparse coding, the grouped atoms happen to be diverse (less correlated). Thus, if the given dictionary comprises of correlated atoms those are less likely to be grouped together in the SAC approach. This affects the classification performance due to inconsistency in sparse coding. Addressal the above mentioned weakness in the estimated block structure is the prime motivation behind this work. 

Further, in the context of image recognition, we come across a proposal of supervised block-structured dictionary learning approach that employs intra-block coherence suppression for reducing the redundancy and is referred to as the IBCS~\cite{chi2013block} dictionary. The minimization of intra-block coherence in a dictionary is critical for consistency in the resulting sparse codes. In that work, the block structure is initialized in a supervised manner and is kept fixed during the dictionary learning. It would be interesting to explore the adaptation of block structure while retaining the class supervision. Motivated by these works, we propose a classification-driven dictionary learning approach and contrast its performance with existing approaches on synthetic data as well as real data. The main contributions of this work are as follows:
\begin{itemize}
	\item A novel block structuring algorithm is proposed that exploits the similarity among dictionary atoms rather than that among the resultant sparse codes.
    \item The proposed  block formation approach is shown to reduce the \emph{inter-block} coherence as well as providing a more precise control on the block size in contrast to the SAC algorithm.
	\item Use of class supervision in block formation for enhancing the classification performance  achieved with the learned block-structured dictionary.
\end{itemize}   


The remainder of the paper is organized as follows. First, the prior work on the dictionary learning using the block structure is discussed in Section~\ref{sec:PWDD}. The proposed correlation based greedy clustering algorithm is discussed in Section~\ref{sec:CBCA}. In Section~\ref{sec:ProbDef}, we formulate the classification driven dictionary learning approach. Section~\ref{sec:synth} and Section~\ref{sec:exp} present the evaluation of the proposed approach on synthetic and real data, respectively. The paper is concluded in Section~\ref{sec:Conclusion}.

\section{Prior work on block-structured dictionary}
\label{sec:PWDD}
The idea behind learning a block-structured dictionary is to exploit any structure that is embedded in the signals for producing more efficient sparse representation. A variety of algorithms have been proposed in the literature for this purpose. In initial works~\cite{stojnic2008reconstruction,eldar2009robust}, it is assumed that the block structure of the dictionary is known a priori.  Later, in~\cite{zelnik2012dictionary}, an unsupervised SAC algorithm is proposed for deriving the block structure from the data. The dictionary is learned using the BKSVD algorithm while iteratively updating both the block structure and the atoms of the dictionary. Learning of a block-structured dictionary $\bm{D}$ along with its block structure $\bm{b} \in R^{n_a}$ having a maximum block size of $b_s$ can be formulated as,
\begin{eqnarray}\label{basicBKSVD_obj}
&\underset{\bm{D,b,U}}{\operatorname{min}} \quad&\norm{\bm{Y-DU}}_{F}  \nonumber\\
&\text{such that}\quad&\norm{\bm{u}_{i}}_{0,\bm{b}} \le p, \ i = 1,\ldots,n_s \nonumber\\
&&|\bm{b}_{j}| \leq b_{s}, \ j= 1,\ldots,n_b
\end{eqnarray}
where $\bm{D} =[\bm{d}_1,\bm{d}_2,\ldots,\bm{d}_{n_a}]$ is the dictionary having $n_a$ numbers of $m$-dimensional atoms, $\bm{Y}=[\bm{y}_1,\bm{y}_2,\ldots,\bm{y}_{n_s}]$ is the data matrix, $\bm{U}=[\bm{u}_1,\bm{u}_2,\ldots,\bm{u}_{n_s}]$ is sparse code matrix, $\norm{.}_{F}$ is the Frobinous norm, $\norm{\bm{u}_{i}}_{0,\bm{b}}$ is the $l_0$-norm over $\bm{b}$ and finds the number of non-zero blocks,  $\bm{b}_{j} = \{i \in 1,\ldots, n_a|\bm{b}[i] = j\}$ is set of indices in the $j$th block, $p$ is chosen block sparsity and $n_b$ is the number of blocks. 

The dictionary update process in the KSVD and BKSVD algorithms is quite similar, except the later involves block-by-block update. In contrast to the KSVD algorithm, the BKSVD algorithm requires about $b_{s}$ times lesser number of SVDs. Thus, the computational complexity in the BKSVD dictionary update is significantly lesser. The SVD ensures that intra-block atoms are orthonormal. This minimizes the redundancy in the dictionary as well as the inconsistency in the sparse coding, hence improving the classification performance.


\subsection{Sparse Agglomerative Clustering}
\label{sec:SAC_review}
\begin{figure}[!t]
	\centering
	\includegraphics[height=4.5cm,width=8.5cm]{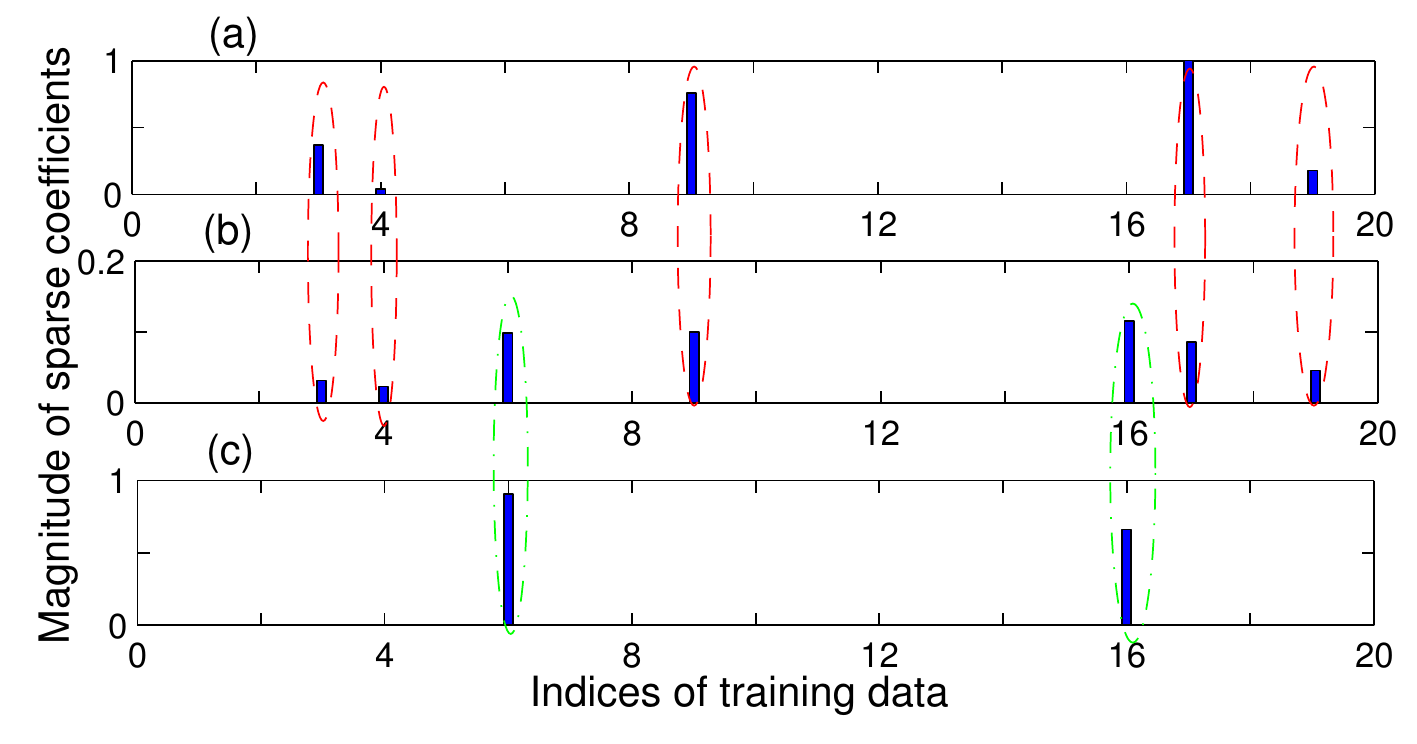}\vspace{-2mm}
	\caption{ \it {Illustration of maximal intersection based group formation in the SAC process. The sparse codes belonging to atoms $\bm{d}_1, \ \bm{d}_2,$ and $\bm{d}_3$ of a $3$-atom toy-dictionary for $20$ randomly selected training vectors are shown in plots (a), (b) and (c), respectively.}}
	\label{fig:ReasonExpla}
\end{figure}

The SAC algorithm employs an iterative process for estimating the block structure $\bm{b}$ from the sparse code matrix $\bm{U}$ of the training data obtained using the OMP. The algorithm starts by considering each atom as a block. At each iteration, it merges two blocks based on the maximum intersection in the involved sparse codes while satisfying the constraint on the maximum block size. To illustrate this merging process, a toy-dictionary has been created by taking three arbitrary atoms $(\bm{d}_1, \bm{d}_2$ and $\bm{d}_3)$ out of a larger sized dictionary $\bm{D}$. Further, $20$ arbitrary selected training data vectors are sparse coded over that toy-dictionary. In this illustration, as the dictionary has only $3$ atoms, there are $3$ possible ways to merge any two atoms to make a block. 
On finding the intersection among the obtained sparse codes, the pair of atoms having the largest intersection among the sparse codes are grouped together. Fig.~\ref{fig:ReasonExpla} shows the block formation at very first iteration of the SAC algorithm. For atoms  $\bm{d}_1$ and $\bm{d}_2$, the match in the indices of sparse codes happens to be maximum, so these atoms form the first group.  


\begin{figure}[!t]
	\vspace{-5mm}
	\centering
	\includegraphics[scale=0.8]{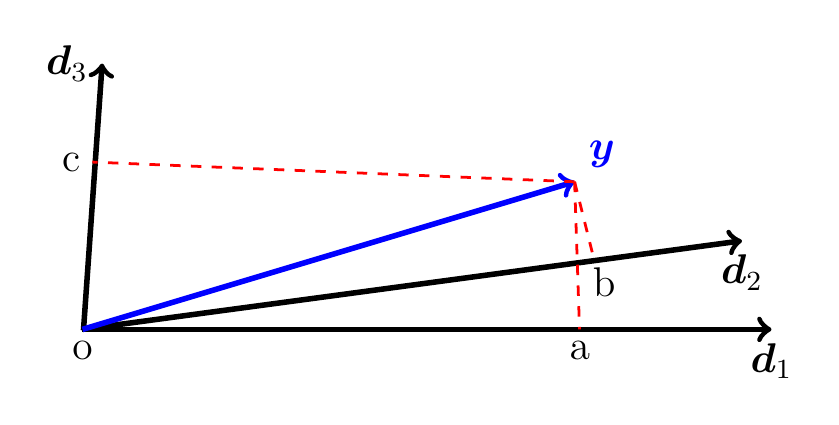}\vspace{-4mm}
	\caption{ \it {Graphical display of the OMP based sparse coding for a $3$-atom dictionary having correlated atoms $\bm{d}_1$ and $\bm{d}_2$.}}\vspace{-5mm}
	\label{fig:OMP_drawback}
\end{figure}

In the existing SAC process, the OMP algorithm has been employed for sparse coding of the data. The OMP, being a greedy iterative algorithm, selects one atom at each step that is most correlated to the current residual. The selected atoms are highly uncorrelated with each other. Thus, in the SAC process, the formed blocks contain diverse atoms rather than similar ones. Assume a dictionary happens to contain two or more moderately correlated atoms. So while sparse coding the data over that dictionary, the OMP algorithm is expected to select any one of them based on the similarity. Fig.~\ref{fig:OMP_drawback} graphically displays the OMP based sparse coding of a target vector $\bm{y}$ over a dictionary having two correlated atoms say $\bm{d}_1$ and $\bm{d}_2$. After selecting either of them, the current residual will no longer lie in the directions of the correlated atoms rather it would become correlated with another atom in the dictionary say $\bm{d}_3$. On account of that there exists a finite possibility that those correlated atoms will appear in different blocks if the SAC process is followed. In classification task, the existing SAC based block-structured dictionary may produce the sparse codes for the same class enrollment and test data involving different blocks. This inconsistency in the sparse coding leads to degradation in the classification performance. One can also employ other sparse coding schemes that do not perform orthogonalization while selecting the atoms unlike the OMP. The least angle regression (LARS)~\cite{efron2004least} algorithm is one such alternative and we hypothesize that it should provide some improvement in the SAC. Following this argument, we modified the existing SAC process to include LARS-based sparse coding and created a new block-structured dictionary. 

\begin{figure}[!t]
	\centering
	\includegraphics[height=3cm,width=7.5cm]{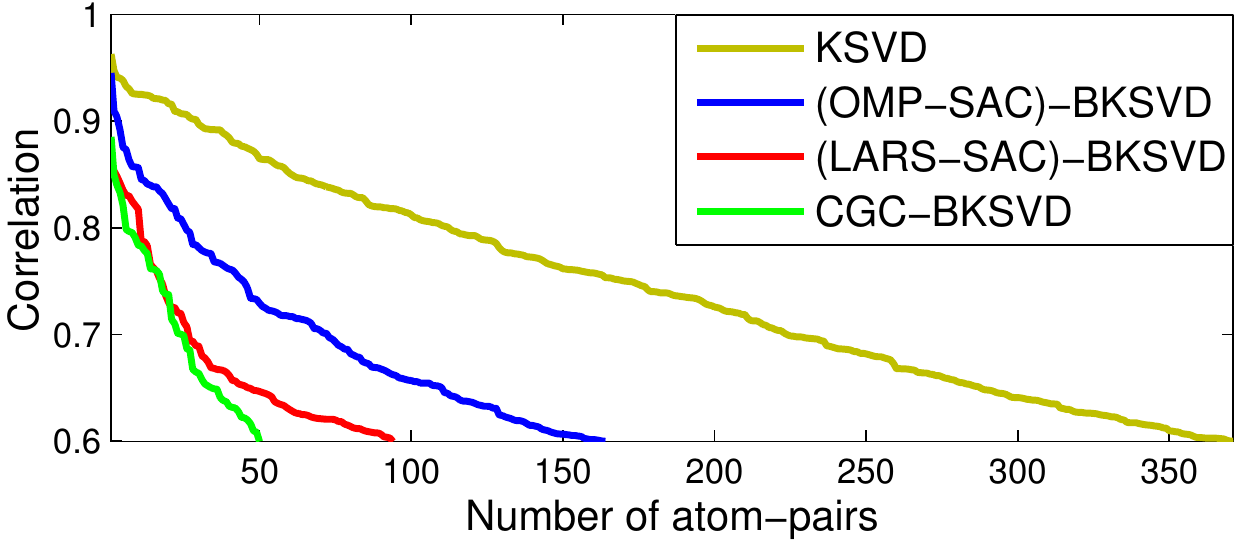}\vspace{-1mm}
	\caption{ \it {The profiles of pairwise correlations (sorted in descending order) among the dictionary atoms for different block-structured dictionaries. It can be seen that use of LARS sparse coding in the SAC has resulted in significant reduction in mutual coherence among dictionary atoms.}}\vspace{-4mm}
	\label{fig:SparkComparison}
\end{figure}

For assessing the quality of the block structure, we have computed the pairwise correlations among all atoms of the OMP-SAC-based and the LARS-SAC-based block dictionaries. Both these dictionaries have been initialized with the same $1200$-atom KSVD dictionary while learning. The number of pairs having correlation value greater than $0.6$ in these two dictionaries are plotted in Fig.~\ref{fig:SparkComparison}. On comparing the number of atom-pairs having higher correlation than the chosen threshold, the LARS-SAC-based BKSVD dictionary is noted to exhibit significantly lower inter-block coherence than the OMP-SAC-based BKSVD dictionary. Later in Section~\ref{sec:res}, we also show that the LARS-SAC-based BKSVD dictionary also yields better SV performance than the existing OMP-SAC-based BKSVD dictionary. Motivated by the improved block structure quality with increased correlation among grouped atoms, we explore the clustering of the atoms into block structure based on correlation criterion rather than the intersection among indices of the sparse codes. In the next section, we describe the correlation based greedy clustering algorithm for producing the block-structured dictionary. Following that we explore the inclusion of class information in block formation to produce more discriminative block-structured dictionary.

\begin{figure*}[!t]
	\centering
	\includegraphics[height=11cm,width=7.0cm]{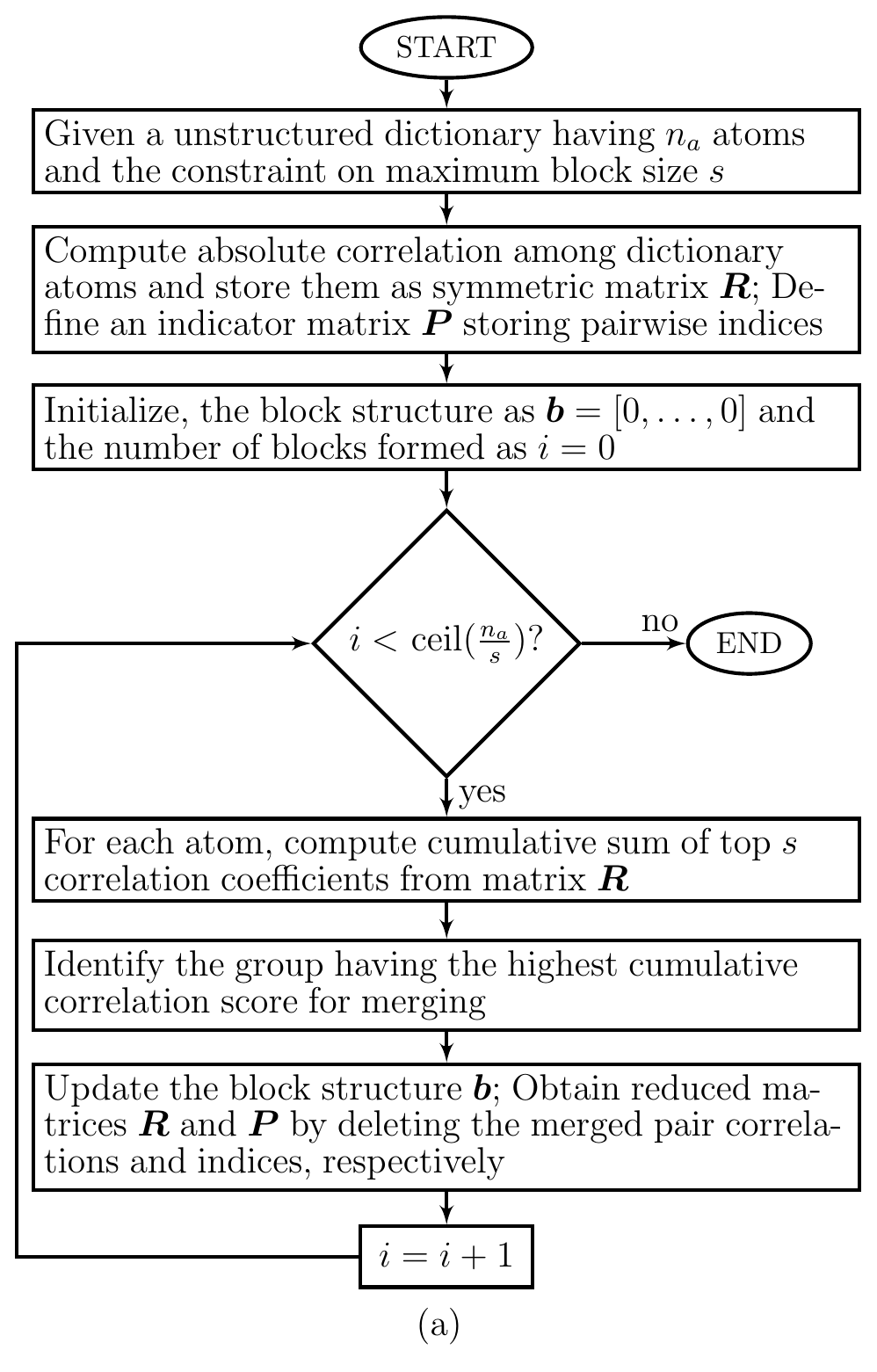}\hspace{1cm}
	\includegraphics[height=10cm,width=8cm]{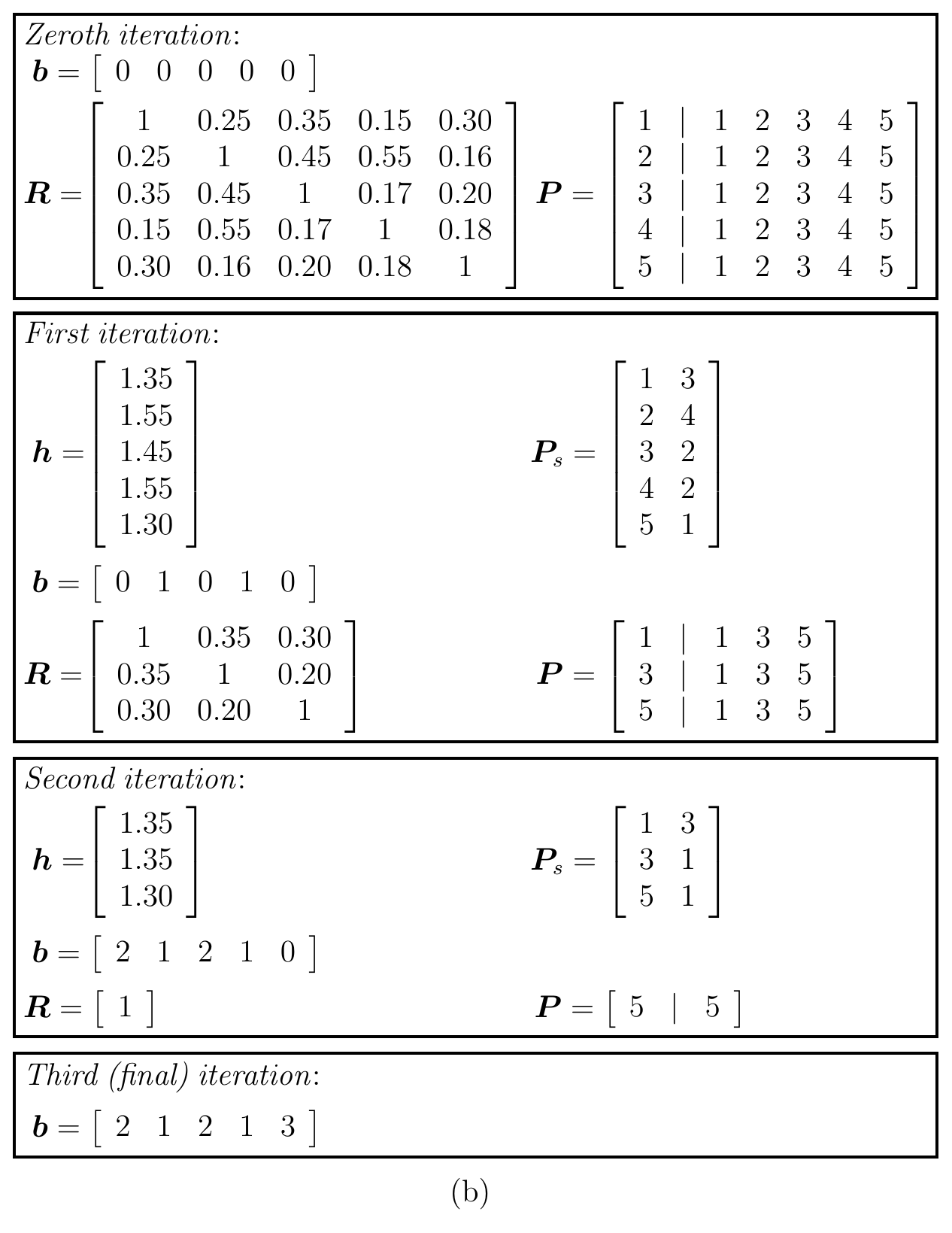}\vspace{-0mm}
	\caption{ \it {The CGC block structuring algorithm (a) flow chart,  (b) an example explaining the involved steps in the context of a dictionary having $5$ atoms and the maximum block size of $2$.}}\vspace{-5mm}
	\label{fig:AlgoExplanation}
\end{figure*}

\section{Correlation based Greedy Clustering Algorithm}
\label{sec:CBCA}

In this section, we propose an approach for determining the block structure $\bm{b}$ given a dictionary $\bm{D}$ exploiting the similarity among its atoms. As the clustering is done on the basis of the pairwise correlation among the dictionary atoms, we refer to this approach as \emph{Correlation} based  \emph{Greedy Clustering} (CGC) algorithm.  The flow diagram of this algorithm is shown in Fig.~\ref{fig:AlgoExplanation}(a) along with an example explaining the involved steps. 

Initially, no block structure is assumed, i.e., $\bm{b}=[0,\ldots,0]$.  From the given dictionary, first the absolute correlations among the atom-pairs are computed and arranged in form of a symmetric matrix $\bm{R}$. For storing the information about the pair indices an indicator matrix  $\bm{P}$ is created. First column of the matrix $\bm{P}$ denotes the indices of dictionary atoms that are yet to be assigned to any block. Each row of $\bm{P}$, excluding the very first entry, simply lists the indices of all the atoms that are available for grouping. At each iteration, the CGC algorithm finds a predefined block size $b_s$ numbers of top most correlated atoms. For this purpose, the cumulative sum of largest $b_s$ correlation values in each row of $\bm{R}$ is computed and stored in a local vector $\bm{h}$.  The indices of selected top correlated atoms are stored into another local selection matrix $\bm{P}_s$. The group of atoms that result in highest cumulative correlation sum form a new block. After finding the new block, first the block structure $\bm{b}$ is updated. Following that the matrices $\bm{R}$ and $\bm{P}$ are updated by discarding the correlation and index information of all those atoms that have formed the block, respectively. These steps are repeated until all dictionary atoms have been assigned to a block. At each iteration only one block is formed and when the maximum block size criterion is no longer satisfied the remaining atoms are group in a lower size block. For fast update, the order of the indices of yet to be grouped atoms in $\bm{P}$ should be kept same as in the symmetric matrix $\bm{R}$ as shown in Fig.~\ref{fig:AlgoExplanation}(b).

In CGC algorithm, with formation of every new block the value of the pairwise correlation for the unassigned atoms keep decreasing. This results in the last few unassigned atoms exhibiting very small correlation. Assigning these atoms to predefined bigger size block may affect the consistency of the sparse coding. To address this issue, we have gradually reduced the maximum block size in steps of one when about $20\%$ of the total atoms of the dictionary are left to be grouped. 

The steps of the CGC algorithm are illustrated in Fig.~\ref{fig:AlgoExplanation}(b) by considering an arbitrary dictionary having $5$ atoms. For ease of illustration, the maximum block size $b_s$ is kept as $2$. At beginning, the dictionary block structure is initialized as $\bm{b} = [0, 0, 0, 0, 0]$. In the first iteration, the atoms having indices  $2$ and  $4$ are found to have the highest correlation. So those  form the first block. In the second iteration, the highest correlation is noted for atoms having indices $1$ and $3$, so the second block is formed by them. In the third iteration, only $5$th atom is left to be assigned, so it alone forms the third block. Thus, the estimated block structure turns out as $\bm{b} = [2, 1, 2, 1, 3]$.

\section{Classification Driven Block-Structured Dictionary}
\label{sec:ProbDef}
In addition to the SAC-BKSVD based dictionary, we find a proposal of intra-block coherence suppression (IBCS)~\cite{chi2013block} based block-structured dictionary in the context of image recognition task. Unlike the SAC-BKSVD approach, in the IBCS dictionary learning, the class labels are used in determining the blocks and so obtained block structure is kept fixed during dictionary learning. On exploring in SV task, we found that the IBCS based block-structured dictionary outperforms both SAC-BKSVD and CGC-BKSVD based dictionaries. Despite having unadapted block structure, the improved detection cost obtained for the IBCS dictionary highlights the impact of the class supervision in block formation. Motivated by that, we propose a novel block structuring scheme that allows the grouping of atoms within-class only.

The proposed scheme is intended to enhance the ability of a dictionary to produce more discriminative sparse codes. The discriminative sparse codes should exhibit following property
\begin{equation}\label{discrimSC}
\bm{U}^{k}_{\bm{b}(c)} = 0, \,\,\forall \,\, k, c \in (1, 2,\ldots,C), \ \ k \neq c
\end{equation}
where  $\bm{U}^{k}_{\bm{b}(c)}$ is the sparse coefficient matrix for the $k$th class training dataset $\bm{Y}^{k}$ and the atoms in block structure $\bm{b}(c)$ corresponding to the $c$th class. Therefore, in ideal case, all non-zero coefficients for $\bm{Y}^{k}$ correspond to $\bm{b}(k)$ only. 

Towards achieving this goal, we define an objective function for learning a discriminative dictionary as
\begin{eqnarray*}
\underset{\bm{D,b,U}}{\min} \, \Big\{\norm{\bm{Y-DU}}_{F}^{2} + 
\sum\limits_{j=1}^{n_b}\Big(\sum_{(r, s) \in I(j),\, r\neq s} \norm{\bm{d}_{r}^{T}\bm{d}_{s}}^{2} \Big)\dots&  \nonumber \\ 
 +\sum\limits_{c=1}^{C}\norm{\bm{Y}^{c} - \bm{D}_{c}\bm{U}^{c}_{\bm{b}(c)}}_{F}^{2} \Big\}\ \quad\quad\quad \nonumber& \\
 \end{eqnarray*}
 \vspace*{-9mm}
 \begin{eqnarray}\label{KSVD_term_obj}
\text{s.t.}\,\,\norm{\bm{u}_{i}}_{0,\bm{b}} \le p,\, |{\bm{b}_{j}}| \le b_s, \, i = 1,\ldots,n_s &
\end{eqnarray}
where $\bm{Y}^{c}$ is the $c$th class training data, $\bm{D}_{c}$ is the $c$th class sub-dictionary, $n_b$ is number of blocks in the dictionary $\bm{D}$ and, $\bm{b}_{j}$ and  $I(j)$ denote the $j$th block and its indices, respectively. In (\ref{KSVD_term_obj}), the first term ensures the good reconstruction ability, the second term reduces the intra-block redundancy and the third term enhances the discrimination in sparse codes to aid the classification. The simultaneous optimization of all three constraints in (\ref{KSVD_term_obj}) may not be feasible. 

Here, we wish to highlight that the first two constraints in (\ref{KSVD_term_obj}) can be optimized by evoking the existing BKSVD dictionary learning technique. In fact, it would ensure that all intra-block coherences are zero. The formation of blocks using either the SAC or the proposed CGC algorithm does not utilize the class information. So the atoms within the blocks may belong to two or more classes. As a result of that, the dominant coefficients in the sparse coding of training data belonging to different classes could involve the same set of blocks. With multi-class data being involved with  a block, the updated dictionary will lose the ability to produce more discriminative sparse coding of the data. Towards  addressing this issue, we have explored the inclusion of the class supervision in the block formation. In the following sub-sections, the details of supervised block structuring and the dictionary update using the well-known SVD approach are presented. The overview for learning the proposed dictionary is given in Algorithm \ref{alg:RDMC}.

\subsection{Block structure: initialization and update}
For including the class supervision in the block formation, first the dictionary is initialized by selecting predefined number of examples from each of the $C$ classes.  Let the class indices for such a dictionary be stored in a vector $\bm{l}$ defined as 
\begin{equation}
\bm{l} = [\underbrace{1,\ldots,1}_{|\bm{l}_{1}|},\ldots, \underbrace{i,\ldots,i}_{|\bm{l}_{i}|},\ldots,\underbrace{C,\ldots,C}_{|\bm{l}_{C}|}]
\end{equation}
where $|\bm{l}_{i}|$ is number of examples in the $i$th class.

Now for optimizing the block structure within each class, the proposed CGC algorithm is invoked in a constrained manner. To preserve the class supervision, appropriate constraints are introduced in the merging process of CGC algorithm to allow the grouping of atoms from the same class only. For this, a simple approach would be to perform block structuring separately for each class while indexing the blocks across the classes uniquely.

\begin{algorithm}[!t]
	\caption{\it Procedure for learning improved block-structured dictionary.}
	\label{alg:RDMC}
	\begin{algorithmic}
		\begin{small}
			\STATE {\bf Input:} Given training dataset $[\mathbf{Y}^1,\ldots,\mathbf{Y}^{C}]$ with labels and maximum number of dictionary atoms $M$ per class.
			\STATE {\bf Step 1.} Obtain the initial dictionary and their corresponding class label $\bm{l}$.
			\STATE {\bf for} fixed number of iterations i.
			\STATE {\bf for} fixed number of iterations j $\leq$ i.
			\STATE {\bf Step 2.} Compute the correlation among dictionary atoms.
			\STATE {\bf Step 3.} Find the block structure $\bm{b}$ using either CGC algorithm or supervised CGC algorithm.
			\STATE {\bf end for}
			\STATE {\bf Step 4.} Compute the sparse coefficients matrix $\bm{U}$ using BOMP algorithm.
			\STATE {\bf Step 5.} Update the block-structured dictionary $\bm{D}$.
			\STATE {\bf end for}
			\STATE {\bf Output:} Dictionaries  $\bm{D}$ and corresponding block structure $\bm{b}$.
		\end{small}
	\end{algorithmic}
\end{algorithm}

\subsection{Dictionary Update}
\label{sec:dict_update}
Given the block structure and the initial dictionary, the training data is sparse coded using the block-OMP (BOMP) algorithm~\cite{eldar2010block}. For updating the dictionary, all training data vectors associated with each of the blocks in the resulting sparse codes are collected. Let  $\omega_{i}(c)$ denotes the list of indices of all those training data that have nonzero sparse coefficient for the $i$th block in the $c$th class $\bm{b}(c)$. For updating the $i$th block in $\bm{b}(c)$, the representation error excluding the contribution of that block is computed as 
\begin{equation}\label{basicSVD}
\bm{E}^{\omega_{i}(c)} = \bm{Y}^{\omega_{i}(c)} - \sum_{m=[1,C \, ]}\Big(\sum_{\begin{footnotesize} \begin{array}{c} j = [1,n_b(m) ], \\ j \neq i \ \text{if} \ m=c  \end{array}\end{footnotesize}}\bm{D}_{\bm{b}_{j}(m)}\bm{U}^{\omega_{i}(m)}_{\bm{b}_{j}(m)} \Big)
\end{equation} 
where $\bm{E}^{\omega_{i}(c)}$ is the error matrix or the data for the $i$th block in the $c$th class, $n_b(m)$ is the number of blocks in the $m$th class, $\bm{b}_{j}(m)$ is the indices of $j$th block in the $m$th class and remaining terms have the usual meaning. Now $\bm{E}^{\omega_{i}(c)}$ is factorized into $\bm{U}\bm{\Delta}\bm{V}^T$ using the SVD algorithm. The representation error is minimized by replacing the dictionary atoms and the selected sparse coefficients with top $|\bm{b}_{i}(c)|$ rank components obtained using SVD as
\[\bm{D}_{\bm{b}_{i}(c)} = [\bm{U}_{1},\ldots,\bm{U}_{|\bm{b}_{i}(c)|}]\]
and
\begin{equation}\label{BKSVD_dict_update}
\ \bm{U}^{\bm{b}_{i}(c)}_{\omega_{i}(c)} = [\bm{\Delta}_{1}^{1}\bm{V}_{1},\ldots,\bm{\Delta}_{|\bm{b}_{i}(c)|}^{|\bm{b}_{i}(c)|}\bm{V}_{|\bm{b}_{i}(c)|}]^T
\end{equation}

Both dictionary and its block structure are updated to achieve convergence or for pre-determined number of iterations. Obviously, the atoms in the updated $\bm{D}_{\bm{b}_{i}}(c)$  happen to be orthonormal to each other. Thus one of the criteria laid in (\ref{KSVD_term_obj}) is met perfectly.

\begin{figure}[t]
	\centering
	\includegraphics[height=3.5cm,width=8.0cm]{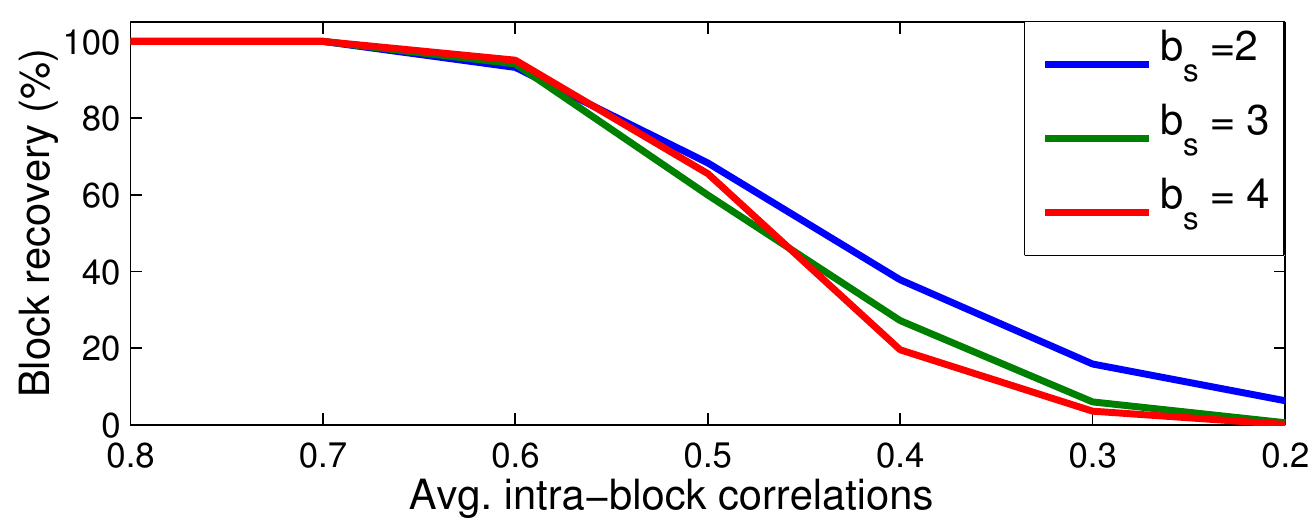}\vspace{-1mm}
	\caption{ \it{Percentage of (exact) block recovery using proposed CGC algorihtm while varing the block size $b_s$ and average intra-block correlations.}}\vspace{-5mm}
	\label{fig:VaringBlkSize_n_Intra_Corr}
\end{figure}

\begin{figure*}[t]
	\centering
	\includegraphics[scale=0.8]{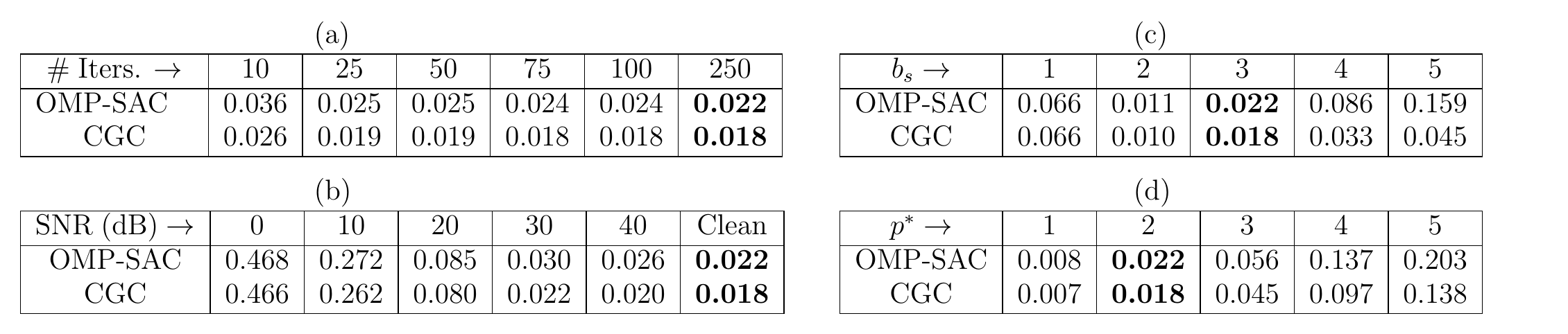}\vspace{-1mm}
	\caption{ \it{Comparison of the reconstruction errors for synthetic data trained block dictionary using OMP-SAC and CGC algorithms under varying conditions: (a) number of iterations involved in learning the dictionary,  (b) signal-to-noise ratio (SNR) of the training data, (c) block size $b_s$ in the learned block structure,  and (d) number of blocks $p^*$ employed in generating the synthetic training data. The highlighted entries across the tables correspond to usage of the identical values of studied parameters.}}\vspace{-5mm}
	\label{fig:CGC_ReconsError}
\end{figure*}

\section{Experiments on the synthetic data}
\label{sec:synth}
In this section, we evaluate the proposed CGC algorithm for recovery of the underlying block structure and compare the reconstruction errors for the KSVD based block dictionaries learned using the OMP-SAC- and CGC-based block structuring algorithms on synthetic data. All experiments are repeated $50$ times and their averaged performance are reported.

\subsection{Block recovery}
\label{sec:blkRecovery}
For this study, we decided to create block dictionaries having $60$ atoms with $30$ dimensional data. For this purpose, first a randomly initialized matrix having $30$ rows and $(60/b_s)$ columns was created, where $b_s$ is chosen block size. For each of the columns in the created matrix, additional $(b_s-1)$ clones were derived by adding random noise in varying scales.  On collecting all these columns, we obtained a $60$-atom initial dictionary $\bm{D}^*$ having an oracle block structure $\bm{d}^*$ such that each block contains $b_s$ atoms. For studying the effect of degree of correlation among the initial dictionary atoms, separate dictionaries were created following the above outline procedure with the average intra-block correlation being controlled by noise added during cloning. For all such synthetically created initial dictionaries, the average inter-block correlation for top $20$ atom-pairs is found to lie in the range $0.51$-$0.57$.

Fig.~\ref{fig:VaringBlkSize_n_Intra_Corr} shows the block recovery performances of the CGC algorithm for varying intra-block correlation and block size. A block is considered to be recovered only when the estimated block indices are identical to those in the oracle block structure $\bm{d}^*$. The CGC algorithm is noted to recover underlying block structure perfectly for average intra-block correlation being $0.7$ or higher regardless the block size considered. Whereas, when the average intra-block correlation drops lower than $0.6$, the accuracy of the block recovery drops sharply owing to decreasing gap between intra- and inter-block correlations. One might wonder such high correlations may not exist in real (speech) data created dictionary, it is already shown in  Fig.~\ref{fig:SparkComparison} that do.

\subsection{Reconstruction performance}
\label{sec:reconsPerformance}
We now evaluate the CGC-BKSVD dictionary in terms of reconstruction performances under varying condition while contrasting with existing OMP-SAC-BKSVD dictionary. The reconstruction performance for the synthetic data matrix $\bm{Y}$ over a learned dictionary $\bm{D}$ is computed through sparse coding with a block sparsity of $3$ and defined as $\frac{\norm{\bm{Y} - \bm{D}\bm{U}}_F}{\norm{\bm{Y}}_F}$.

For generating the data for dictionary learning, we have first created a $60$-atom dictionary using the procedure outlined in \ref{sec:blkRecovery} using synthetic data. For this dictionary, the block size is kept as $3$ and the average intra-block correlation is kept as $0.68$. From so created dictionary having known block structure, weighted sum of $2$ randomly selected blocks is computed to generate a synthetic data vector. Following this scheme $5000$ data samples are derived for experimentation. For assessing the robustness of the dictionary learning approaches, the noisy versions of synthetic data are also created by adding white Gaussian noise at different signal-to-noise ratios (SNRs).

For learning both kinds OMP-SAC- and CGC-based block-structured dictionaries, the same KSVD learned dictionary is used as initialization. Unlike the former which iteratively updated both dictionary and the block structure, in the later only dictionary is updated iteratively while keeping the block structure estimated in first iteration fixed. Therefore, we studied the effect of iterations involved in dictionary learning in two cases. The reconstruction errors for this study are tabulated in Fig.~\ref{fig:CGC_ReconsError}(a). The CGC approach not only converges faster but also outperforms the contrast. 

Fig.~\ref{fig:CGC_ReconsError}(b) shows the impact of addition of noise in the dictionary learning data for OMP-SAC- and CGC-based block dictionaries. The table lists the reconstruction performance for the noiseless data with respect to dictionaries learned using noisy data. Note that, the reconstruction error for the CGC case at $30$ dB SNR matches with that for the OMP-SAC case under noiseless data. Thus, the CGC approach also maintains the edge over the OMP-SAC approach under noisy data. 

In the CGC-based approach, the initial dictionary is clustered based on the pairwise correlation among its atoms. Thus, we hypothesize that the tuning of the block size during the dictionary learning is less critical than that in the OMP-SAC-based approach. Fig.~\ref{fig:CGC_ReconsError}(c) lists the reconstruction errors for varying block size employed in dictionary learning and these results support our hypothesis. 

In the earlier discussed synthetic data generation process, the variability of generated data depends on the number of blocks selected from the oracle block dictionary. For assessing the modeling ability of the proposed approach, different variability data sets are created by varying the number of blocks employed while generating the synthetic data from $1$ to $5$. Separate dictionaries are learned using both the approaches on those data sets and the corresponding reconstruction errors are tabulated in Fig.~\ref{fig:CGC_ReconsError}(d). Even for higher variability data the CGC-based approach is noted to yield better reconstruction performance than that of OMP-SAC-based approach.


\section{Evaluation of Classification Performance}
\label{sec:exp}
In this section, we evaluate the impact of the proposed innovations in the context of speaker verification (SV) task. 
The different SV systems developed in this work are evaluated on telephone condition test data sets in the NIST 2012 SRE~\cite{NIST2012}. Different sparse representation based SV (SR-SV) systems using unsupervised (KSVD, (OMP/LARS)-SAC-BKSVD) and supervised (IBCS and BKSVD) learned dictionaries are developed. The i-vector~\cite{Dehak2011} Gaussian probabilistic linear discriminant analysis (GPLDA)~\cite{Garcia_PLDA} based SV system is also created for primary contrast.
\subsection{Experimental Setup}
The setup employed for the real data experiments is identical to our earlier work~\cite{haris2015robust}. In the following, we briefly mention the essential detail only. For more detailed description about database, performance measure, and signal processing the reader is referred to \cite{haris2015robust}.
The SRE12 speech data set contains a total of $1918$ speakers (female and  male). 
The test set contains $25,698$ telephone recorded speech utterances, from which about $1.18$ million verification trials are created. The test data is partitioned into three subsets based on the environment and noise conditions. For the development of SV system, the speaker's utterances from NIST SRE06, SRE08, and SRE10 data sets have been used. 
From the development data, a total of $16k$ female and $11k$ male data i-vectors are derived. 
The speech data is analyzed to compute commonly used $13$ dimensional mel frequency cepstral coefficients. These are then augmented with their delta and double delta coefficients, thus resulting in a $39$-dimensional final feature vectors. A $1024$ Gaussian mixture based  universal background model (UBM)~\cite{Reynolds2000} is used in gender-dependent modeling.
For evaluating the performance of different SV systems developed, the BOSARIS toolkit\cite{bosaris} has been used. The performance of the SV systems are measured using the detection cost function $\text{C}_{DET}$ as per NIST protocol~\cite{NIST2012}. It is defined as the mean of the normalized detection costs corresponding to the probability of target being set as $0.01$ and $0.001$ respectively. For being evaluated at low false alarm rates, the $C_{DET}$ measure suits for high security applications. The $\textit{cosine distance scoring}$ (CDS) measure is used to find the scores for the SR-SV methods.

For developing the SV system, the telephone recorded development data is partitioned into two parts: train and test. The initial system for tuning  the parameters is trained on the $\textit{dev-train}$ dataset and then evaluated on the $\textit{dev-test}$ dataset. A total of $40k$ dev-trials are created from $\textit{dev-test}$ data containing $1100$ female and $725$ male speakers. All SV systems explored in this work are modeled in gender-dependent manner.

\subsubsection{Factor Analysis based Modeling}
Gender-dependent UBMs are learned using the telephone speech data derived from $1100$ female and $725$ male speakers. The utterances in the development data are redistributed to have an average duration of $150$ seconds after voice activity detector for being of varying duration. In the i-vector based contrast SV system, $600$ dimensional representational vectors are derived using the \emph{total variability} matrix (T-matrix)  learned on the telephone recorded data. In GPLDA modeling, $500$ dimensional speaker subspaces are used and it includes whitening, length normalization followed by projection into the unit sphere. The GPLDA parameters are learned using pooled telephone and microphone recorded development data i-vectors. In the SR-SV systems,  a gender-dependent joint factor analysis (JFA)~\cite{kenny2007speaker} is employed for session/channel compensation of the Gaussian mixture model (GMM)-UBM mean supervectors.  Following that $300$ dimensional \emph{speaker factors} are derived and further details of the same are available in our earlier work~\cite{kumar2016class}. 


\renewcommand{\arraystretch}{1.25}
\begin{table*}[!t]
	\centering
	\caption{ \it{Performances of the proposed block-structured dictionary based SV systems and those of contrast systems on the NIST SRE12 telephone recorded test data set.}}
	\begin{tabular}{|c|c|c|c|c|c|c|c|c|c|c|c|}\hline
		\multicolumn{3}{|c|}{SV System} & \multicolumn{3}{c|}{Block Structure}     & System  & \multicolumn{4}{c|}{Detection Cost ($\text{C}_{DET}$)} & \%EER  \\ \cline{1-6} \cline{8-12}
		& Dictionary type & Feature/Classifier	& Size & Class Sup. & Updation & Code &  TC2 & TC4 & TC5 & \textbf{Avg.} & \textbf{Avg.} \\ \hline \hline
		
		\multirow{7}{*}{\begin{sideways}Contrast\end{sideways}} &  T-matrix & i-vector/Bayes & - & - & - & $\text{S0}$  & 0.411  & 0.543 & 0.446 & \textbf{0.467} & \textbf{5.22} \\  \cline{2-12}
		
		& JFA-matrix & spk-factor/CDS & - & - & - & $\text{S1}$          & 0.523  & 0.634  & 0.552 & \textbf{0.570} & \textbf{5.72} \\ \cline{2-12} 
		& KSVD & \multirow{5}{*}{ sparse-vector/CDS}	 & - & - & - & $\text{S2}$          & 0.494  & 0.605  & 0.516 & \textbf{0.538} & \textbf{9.94} \\ \cline{2-2} \cline{4-12}
		& OMP-SAC-BKSVD & 	 & variable & no & adapted & $\text{S3}$ & 0.447  & 0.527  & 0.427 & \textbf{0.467} & \textbf{14.40} \\ \cline{2-2} \cline{4-12}
		& LARS-SAC-BKSVD  & 		& variable & no & adapted &        $\text{S4}$              & 0.428  & 0.508  & 0.430 & \textbf{0.455} & \textbf{11.31} \\ \cline{2-2} \cline{4-12}
		& IBCS   & 			& variable& yes& unadapted & $\text{S5}$                    & 0.450  & 0.475  & 0.399 & \textbf{0.441} & \textbf{17.72} \\ \cline{2-2} \cline{4-12}
		& Sup. block-BKSVD& 		& variable& yes& unadapted &  $\text{S6}$        & 0.438  & 0.497  & 0.408 & \textbf{0.447} & \textbf{13.22} \\ \cline{1-1} \cline{2-2} \hline\hline
		
		\multirow{3}{*}{\begin{sideways}Proposed\end{sideways}}       & \multirow{2}{6em}{CGC-BKSVD} &  \multirow{3}{*}{ sparse-vector/CDS}	 & fixed& no& adapted  &  $\text{S7}$  & 0.424  & 0.511  & 0.410 & \textbf{0.448} & \textbf{12.23} \\  \cline{4-12}
		&       & 	& variable& no& adapted & $\text{S8}$             & 0.422  & 0.502  & 0.403 & \textbf{0.442} & \textbf{12.25} \\   \cline{2-2} \cline{4-12}
		&  Sup. CGC-BKSVD &		& variable& yes& adapted & $\text{S9}$    & 0.386  & 0.443  & 0.366 & \textbf{0.398} & \textbf{13.51} \\ \hline\hline   
		
		\multicolumn{6}{|c|}{\multirow{2}{*}{Fusion of systems}} & $\text{S0}$+$\text{S5}$   & 0.362  & 0.434  & 0.364 & \textbf{0.387} & \textbf{4.34} \\   \cline{7-12}
		\multicolumn{6}{|c|}{\multirow{2}{*}}          & $\text{S0}$+$\text{S9}$   & 0.313  & 0.397  & 0.336 & \textbf{0.349} & \textbf{4.14} \\ \hline	    
	\end{tabular}\vspace{-2mm}
	\label{TCWiseRes1}
\end{table*}
\renewcommand{\arraystretch}{1}

\subsubsection{Sparse Representation based Modeling}
The gender-dependent KSVD dictionaries are randomly initialized by selecting development data corresponding to $1200$ and $800$ utterances for female and male cases, respectively. For learning KSVD dictionaries, $50$ iterations are performed. The unsupervised BKSVD dictionaries are initialized with corresponding KSVD dictionaries. Whereas, each of the supervised dictionaries is initialized with class-specific KSVD learned sub-dictionaries having a maximum of $6$ atoms per class (speaker). All kinds of dictionaries are trained on the speaker factors pooled from both the telephone and the microphone development data. Learning of the IBCS and SAC-BKSVD based dictionaries usually require $25$ iterations, while the CGC-BKSVD and supervised CGC-BKSVD based dictionaries is noted to converge in $5$ iterations. All the block-structured dictionary are learned keeping the block size and block sparsity of $3$ and $4$, respectively. In all SR-SV systems, the sparse coding of the enrollment and test data is done using the BOMP algorithm. The coding over the unsupervised and the supervised dictionaries employ the sparsity value of $50$ and $30$, respectively.

\subsection{Results and Discussions }
\label{sec:res}
In this subsection, first the performances of the proposed block structuring approach and class-supervised block-structured dictionary based SV systems are discussed. Following that, the robustness of proposed SR-SV system is evaluated. Finally, the results of the fusion of the proposed SR based and the i-vector based SV systems are presented.


\subsubsection{SV performance evaluation}
The system performances are primarily evaluated in terms of $\text{C}_{DET}$ and the corresponding equal error rates (EERs) are reported only for reference purposes. The three conditions in the NIST SRE12 telephone test data sets are referred to as TC2, TC4 and TC5. The performances of different proposed and contrast SV systems are presented in Table~\ref{TCWiseRes1}. On comparing between the i-vector GPLDA (S0 system) and the KSVD dictionary (S2 system) based SV approaches, we note that the former significantly outperforms the later. As the KSVD dictionary is learned without any supervision or block structure, a direct comparison between the S0 and S2 systems may not be fair. For this purpose, we also did CDS directly on speaker factors and the resulting SV approach (S1 system) is found to be inferior to the S2 system in term of $C_{DET}$. In an earlier work, it is shown that the block-structured KSVD dictionary outperforms the simple KSVD dictionary in the context of SR-SV~\cite{sreeram2015improved}. The remaining performances given in Table~\ref{TCWiseRes1} are discussed next in the context of two enhancement proposed for learning the block-structured dictionary. 

{\it Modified block structuring approach:}
In Section~\ref{sec:SAC_review}, it is shown that LARS-SAC-based dictionary has reduced inter-block correlations, thus it is expected to yield improved SR-SV performance. From Table~\ref{TCWiseRes1}, we note that the LARS-SAC-BKSVD dictionary (S4 system) results in a relative improvement of $2.57\%$ in  $C_{DET}$ and $21.46\%$ in EER over the OMP-SAC-BKSVD dictionary (S3 system). With fixed size block formation, the CGC-BKSVD dictionary (S7 system) is noted to yield in a relative improvement of  $4.07\%$  and  $1.54\%$ in  $C_{DET}$ when compared with the S3 and S4 systems, respectively. Further, on allowing variable block sizes in the CGC-BKSVD dictionary (S8 system) an additional relative improvement of $1.34\%$ in  $C_{DET}$ is obtained.\\  
{\it Supervision in the block formation:}
In Table~\ref{TCWiseRes1}, the S5 system refers to the evaluation of recently proposed IBCS dictionary learning approach in SV task. In IBCS dictionary learning, a class-supervised block structure is employed and the intra-block coherences are minimized using gradient approach without updating the block structure. In contrast to that all previously discussed block dictionaries are learned using SVD along with updating the block structure. Therefore, for direct contrast with the IBCS dictionary, it would be interesting to explore the impact of employing the supervised block structure in the SVD based block dictionary learning. For this purpose, we have learned a block dictionary using BKSVD algorithm but with a class-supervised block structure which is not updated during learning. The resulting block dictionary based SV system is referred to as S6 system. It can be seen from Table~\ref{TCWiseRes1} that both S5 and S6 systems yield similar SV performances. These results demonstrate the impact of class supervision in block structure in the dictionary learning. The second proposal of combining class supervision in the CGC approach for deriving the block-structured dictionary is referred to as S9 system. It can be noted that the S9 system has consistently outperformed previously discussed system on all three test sets in the primary measure $C_{DET}$.

\begin{figure}[!t]
	\centering
	\includegraphics[height=3cm,width=8.5cm]{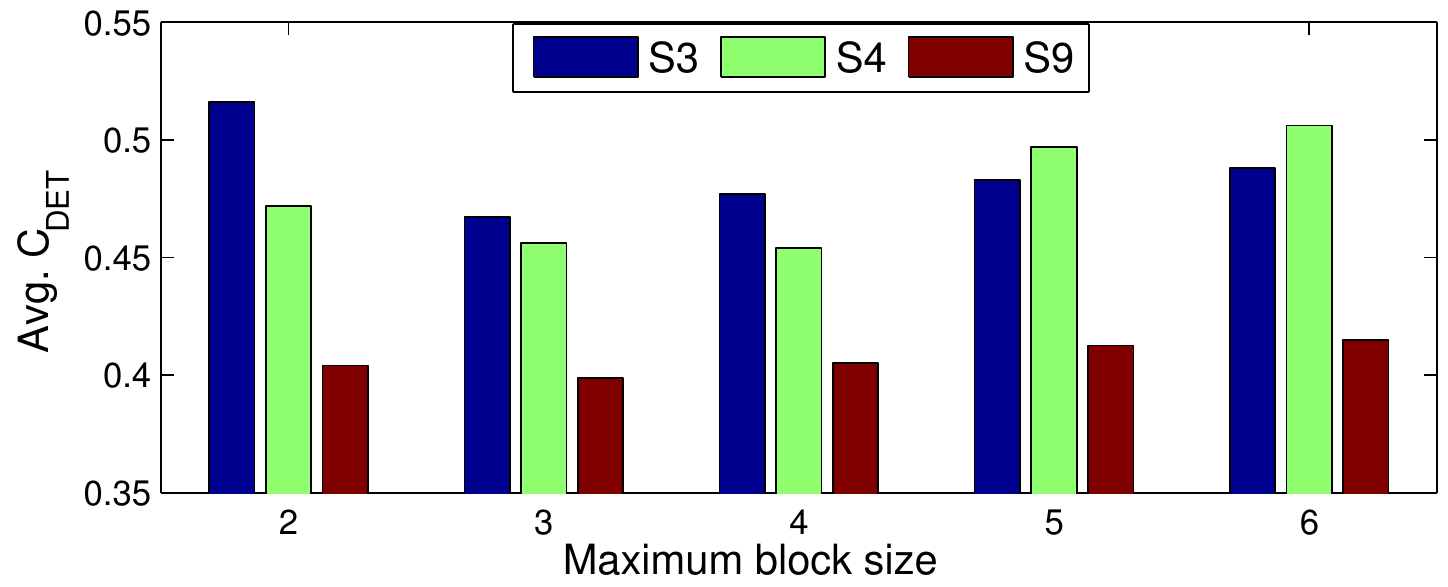}\vspace{-0mm}
	\caption{ \it {Test condition averaged detection performance of the proposed and existing block-structured dictionary based SR-SV systems for varying maximum block size. The porposed (S9) system exhibits lower sensitivity than existing ones.}}
	\label{fig:BlockRobustNess}
\end{figure}

\subsubsection{Sensitivity to maximum block size}
In the previous section, the SV performances of all block-structured dictionary based systems correspond to maximum block size of $3$. We have explored varying the same from $2$-$6$ in the context of three systems (S3, S4, and S9) and the corresponding test condition averaged detection costs are given in Fig.~\ref{fig:BlockRobustNess}. For the chosen block size range, the relative deviations in averaged $C_{DET}$ have turned out to be  $9.56\%$, $10.39\%$ and $3.92\%$ for S3, S4 and S9 systems, respectively. From these results, it can be inferred that the supervised CGC-BKSVD approach is more robust to variation in the block size. For greedy selection being employed in the CGC algorithm and block update using SVD, after a few iterations intra-class atoms become nearly uncorrelated irrespective of the constraint of the block size. This could be the possible reason behind the low sensitivity being exhibited by the proposed approach.

 \begin{figure}[!t]
 	\centering
 	\includegraphics[width=8.5cm,height=6cm]{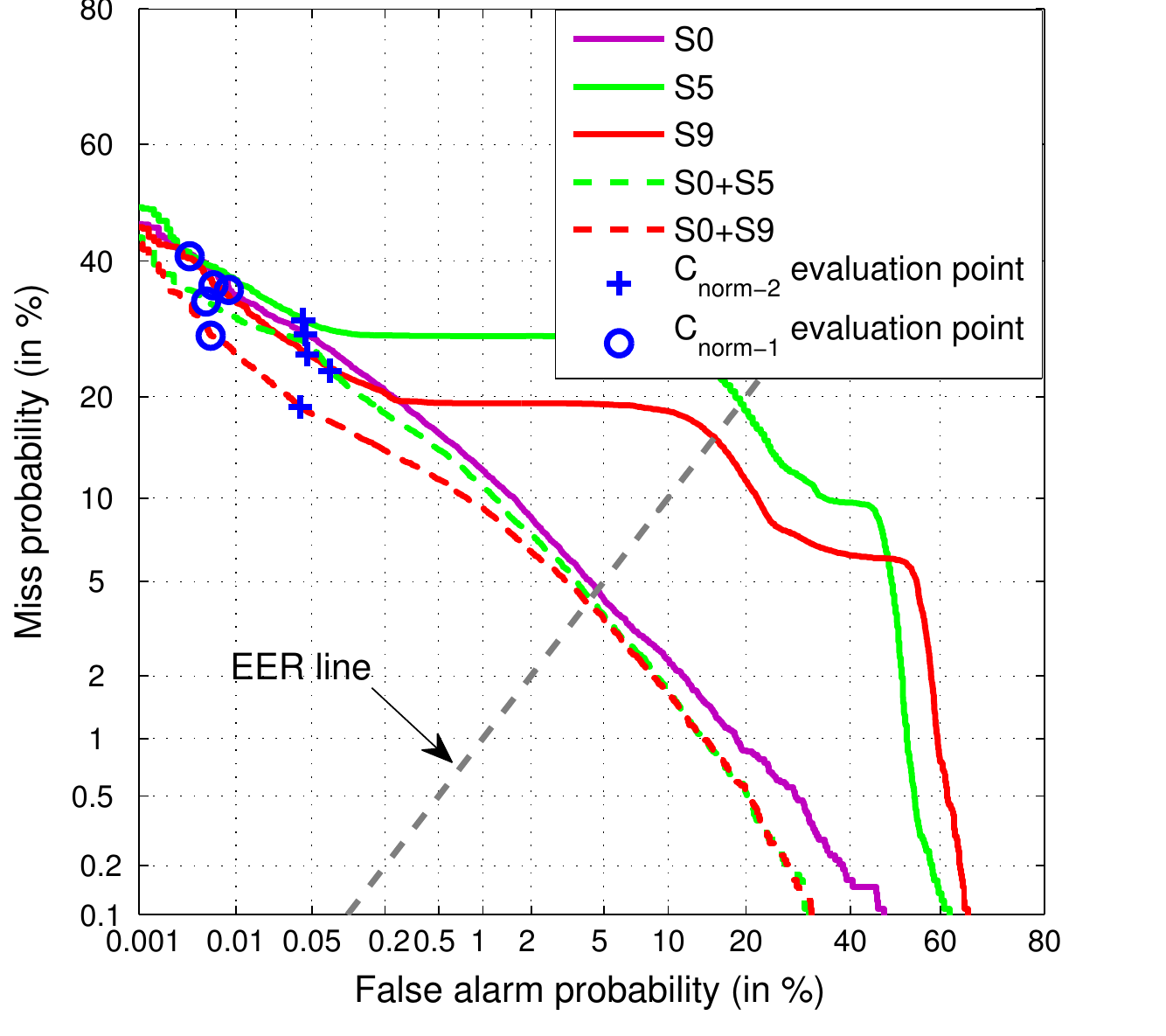}
 	\caption{\it {DET plots for salient SV systems developed in this work on TC2 test condition (S0: i-vector, S5: IBCS, S9: superivsed CGC-BKSVD). Also shown are the fusion of the SR-SV systems with the i-vector SV system.}}\vspace{-0mm}
 	\label{fig:TC2_all_DET_plot}
 \end{figure}

\subsubsection{Exploiting system diversity}
The various SV systems explored in this work mainly differ in terms of the criterion employed in dictionary learning and scoring. 
More specifically, the i-vector based SV approach involves factor analysis for learning the T-matrix and the GPLDA for scoring. Whereas the SR-SV approaches involve cluster wise eigen-decomposition for learning the dictionaries while use the CDS for scoring. To highlight the complementary behavior of these approaches, the DET curves for a few salient systems are plotted in Fig.~\ref{fig:TC2_all_DET_plot}. For exploiting the diversity, the logistic regression based score-level fusion of the i-vector system (S0)  with two best SR-SV systems (S5 and S9) are explored and also shown in Table~\ref{TCWiseRes1}. The best performing fusion (S0+S9) is noted to provide a relative improvements of $12.31\%$ and $20.69\%$ in terms of $C_{DET}$ and EER, respectively, when compared with the individual best performances.

\section{Conclusion}
\label{sec:Conclusion}
In this paper, a novel correlation based block formation approach, referred to as the CGC, is presented for learning block-structured dictionary. The CGC-based block dictionary yields improved reconstruction and classification performances. In contrast to existing SAC-based approach, the proposed one is noted to exhibit faster convergence, lower sensitive to block size and more robust to additive noise while learning the dictionary. For further enhancement in classification ability, the class information is included in the CGC. The resulting block-structured dictionary based SR-SV system provides a relative improvement of $9.75\%$ in the detection cost over the best contrast SR-SV system employing an existing supervised block-structured dictionary. On fusing the best proposed SR-SV and the state-of-the-art i-vector SV systems significant improvements in the classification performance are noted both in terms of the detection cost and the equal error rate.

\appendices
\bibliographystyle{IEEEtran}
\bibliography{reference_all}

\end{document}